\documentclass[preprint,12pt]{elsarticle}

\usepackage{amssymb}
\usepackage{amsmath}
\usepackage{url}

\journal{Knowledge-Based Systems}

\begin{document}

\begin{frontmatter}



%\title{Exploiting Pre-trained Sequence-to-Sequence Transformers for Constituent Parsing}
\title{Exploiting Pre-trained Encoder-Decoder Transformers for Sequence-to-Sequence Constituent Parsing}


\author{Daniel Fernández-González\corref{cor1}} %% Author name
\cortext[cor1]{Corresponding author.}
\ead{danifg@uvigo.gal}

\author{Cristina Outeiriño Cid} 
\ead{cristina.outeirino@uvigo.gal}
\affiliation[label1]{organization={Universidade de Vigo, Departamento de Informática},%Department and Organization
            addressline={Campus As Lagoas s/n}, 
            city={Ourense},
            postcode={32004}, 
            state={},
            country={Spain}}

%% Abstract
\begin{abstract}
To achieve deep natural language understanding, syntactic constituent parsing plays a crucial role and is widely required by many artificial intelligence systems for processing both text and speech. A recent approach involves using standard sequence-to-sequence models to handle constituent parsing as a machine translation problem, moving away from traditional task-specific parsers. These models are typically initialized with pre-trained encoder-only language models like BERT or RoBERTa. However, the use of pre-trained encoder-decoder language models for constituency parsing has not been thoroughly explored. To bridge this gap, we extend the sequence-to-sequence framework by investigating parsers built on pre-trained encoder-decoder architectures, including BART, mBART, and T5. We fine-tune them to generate linearized parse trees and extensively evaluate them on different linearization strategies across both continuous treebanks and more complex discontinuous benchmarks. Our results demonstrate that our approach outperforms all prior sequence-to-sequence models and performs competitively with leading task-specific constituent parsers on continuous constituent parsing.
\end{abstract}


%% Keywords

\end{frontmatter}


\section{Introduction}
\label{sec1}
Syntactic structure is essential in language comprehension, supporting deeper understanding and accurate text generation across a wide range of natural language processing (NLP) tasks. Among various syntactic analysis approaches, constituency grammar is particularly effective at capturing the hierarchical phrase structure of sentences and represent them as a constituent tree (as the one depicted in Figure~\ref{fig:linearizations}(a)). This rich structural representation can enhance linguistic analysis and improve performance in downstream tasks such as machine translation \citep{gu-etal-2018-top,wang-etal-2018-tree,currey-heafield-2019-incorporating},
opinion mining \citep{opinionCP}, sentiment classification \citep{yin-etal-2020-sentibert,bai-etal-2021-syntax}, information extraction \citep{jiang-diesner-2019-constituency,baac070,yang-etal-2020-constituency},  coreference resolution \citep{Swabha18}, semantic textual similarity \citep{sym17040486}, semantic role labelling \citep{wang-etal-2019-best}, among others.

Over the past decades, a variety of techniques have been developed for implementing constituent parsers. The last innovation, proposed by \citet{Vinyals2015}, involves leveraging general-purpose sequence-to-sequence models to directly translate input text into phrase structure trees, drawing inspiration from the success of such models in machine translation tasks \citep{Sutskever2014}. This approach performs constituent parsing using a task-agnostic model, eliminating the need for specialized parsing algorithms. Given a sequence of words, the model predicts a corresponding sequence of tokens that linearizes the parse tree. This approach was not only proposed for building continuous constituent trees, but it was recently applied to produce \textit{discontinuous} constituent trees \citep{FERNANDEZGONZALEZ202343} (as the one in Figure~\ref{fig:linearizations}(b)). The latter are necessary to fully describe all linguistic phenomena present in human languages and are considered one of the most complex syntactic formalisms.

However, while recent efforts
on sequence-to-sequence constituent parsing provided promising results, this trend did not reach state-of-the-art results as it did on other natural language processing tasks such as machine translation \citep{liu-etal-2020-multilingual-denoising,seq2seqMTAAAI} or speech recognition \citep{abs-1904-11660,2020arXiv200605474W}.  In fact, they lagged behind classic parsers based on explicit tree-structured algorithms and supported by a more extensive research background such as transition-based algorithms \citep{attachjuxtapose,coavoux2019b} and chart-based methods \citep{mrini-etal-2020-rethinking,Corro2020SpanbasedDC}.

In order to further improve the performance of the sequence-to-sequence approach, we explore the use of pre-trained encoder-decoder language models as the backbone of a constituent parser. So far, sequence-to-sequence models used for constituent parsing were initialized exclusively with pre-trained encoder-only Transformers such as BERT \citep{devlin-etal-2019-bert} or RoBERTa \citep{roberta}. We propose to use language models where both encoder and decoder are pre-trained, including BART \citep{lewis-etal-2020-bart}, T5 \citep{2020t5} and mBART \citep{liu-etal-2020-multilingual-denoising}. As tree linearizations, we design novel variants based on the strategies presented in \cite{fernandez-gonzalez-gomez-rodriguez-2020-enriched} and \cite{FERNANDEZGONZALEZ202343}, and implement different parsers in combination with the available pre-trained language models. By extensively testing these resulting parsers, we demonstrate that the proposed approach achieves state-of-the-art F-scores among sequence-to-sequence models and performs competitively with leading task-specific parsers on the main continuous benchmark, while also delivering strong results on discontinuous treebanks.

Therefore, our main contributions are:
\begin{itemize}
    \item The development of a novel sequence-to-sequence constituent parser based on pre-trained encoder-decoder language models.\footnote{Our system's code will be released after acceptance.}
    %\footnote{Source code available at \url{https://github.com/danifg/Disco-Seq2seq-Parser}.
    \item New linearization methods are proposed to construct both continuous and discontinuous constituent trees, allowing the model to better exploit the capabilities of a pre-trained decoder (an aspect that had not been explored previously because earlier approaches relied on encoder-only pre-trained Transformers).
    %, which was not necessary in previous approaches that relied on pre-trained encoder-only Transformers.
    \item An extensive evaluation with three different pre-trained encoder-decoder language models (in combination with different linearizations) is performed on the widely-used English Penn Treebank (PTB) \citep{marcus93} and on the main discontinuous benchmarks: the Discontinuous English Penn Treebank (DPTB) \citep{evang-kallmeyer-2011-plcfrs}, and the German NEGRA \citep{Skut1997} and TIGER \citep{brants02} treebanks. 
    \item The proposed approach outperforms all previously existing sequence-to-sequence models and achieves results nearly matching the best scores reported to date on PTB, as well as obtaining competitive results on the DPTB, NEGRA, and TIGER treebanks.
\end{itemize}

%Actualizar
The remainder of this article is organized as follows. Section~\ref{sec:related} reviews prior work on modeling and improving constituent parsing as a sequence-to-sequence task. Section~\ref{sec:methodology} describes the proposed linearization strategies and their use in implementing a constituency parser based on pre-trained encoder-decoder models. Section~\ref{sec:experiments} presents an extensive evaluation on continuous and discontinuous treebanks, together with a detailed performance analysis. Finally, Section~\ref{sec:conclusion} concludes the article.

\section{Related work}%Puede ir antes de las conclusiones
\label{sec:related}
Since the introduction of the first sequence-to-sequence model for constituent parsing by \citet{Vinyals2015}, numerous variants have been proposed to enhance its performance. These developments have primarily targeted two areas: improving the attentional sequence-to-sequence neural architecture \citep{Bahdanau2014}, and refining the linearization strategy used to frame constituent parsing as a sequence prediction task.

Focusing on the original architecture by \citet{Bahdanau2014}, which relies on recurrent neural networks, various modifications have introduced deterministic attention mechanisms \citep{Kamigaito2017,Ma2017,LiuS2S17,Liu2018,fernandez-gonzalez-gomez-rodriguez-2020-enriched}. These approaches aim to replace the probabilistic attention mechanism in order to (1) boost accuracy by deterministically selecting input tokens that are critical for parsing, and (2) accelerate decoding by reducing the need to consider the entire input sequence when computing attention weights. A notable advancement in this direction is the Transformer-based model proposed by \citet{Vaswani2017}, which achieves improvements in both accuracy and processing speed.

On the linearization front, \citet{Vinyals2015} proposed representing parse trees top-down using bracketed notation to group constituents and words. While most subsequent work has adopted this approach, some researchers have explored alternative linearizations. Notably, \citet{Ma2017} and \citet{LiuS2S17} introduced transition-based linearizations using action sequences. \citet{Ma2017} employed actions from the bottom-up transition system of \citet{cross-huang-2016-span}, but found it to underperform compared to the original top-down bracketed method under identical evaluation conditions. In contrast, \citet{LiuS2S17} utilized the top-down transition system of \citet{Dyer2016}, demonstrating that, when paired with a deterministic attention strategy, this approach can offer improved accuracy.

However, these findings were challenged by \citet{fernandez-gonzalez-gomez-rodriguez-2020-enriched}, who showed that the bracketed encoding could be mapped to sequences of actions from \citet{Dyer2016}'s system and that, under the same model architecture, this representation outperforms the top-down action sequence proposed by \citet{LiuS2S17}. Furthermore, they introduced a new linearization method based on the in-order transition system from \citet{Liu2017}, which achieved state-of-the-art results among sequence-to-sequence models for constituent parsing.

Recently, \citet{FERNANDEZGONZALEZ202343} introduced two significant innovations that achieved the highest accuracy to date in sequence-to-sequence constituent parsing. First, they proposed new linearization strategies capable of representing discontinuous constituent trees, thereby extending the applicability of sequence-to-sequence models. Second, they developed a robust neural architecture inspired by the work of \citet{fernandez-astudillo-etal-2020-transition}. Specifically, they designed a modified Transformer model in which certain attention heads are explicitly assigned to focus on words involved in transition-based actions. Their approach is then initialized using the pre-trained encoder-only model RoBERTa \citep{roberta}. Although this approach does not implement a transition-based parser directly, it effectively incorporates structural information to enhance parsing performance. Please note that, in our work, no specific modifications of the original Transformer architecture were performed.

\section{Methodology}
\label{sec:methodology}
 In the sequence-to-sequence framework, constituent parsing is framed as a \textit{translation} task: given an input sentence of $n$ words $\mathbf{w}=w_0, \dots, w_{n-1}$, the goal is to generate a linearized representation of a constituent tree $C$, encoded as a sequence of $m$ tokens $\mathbf{y}=y_0, \dots, y_{m-1}$, where $n < m$. This transformation must be invertible to ensure that the original tree can be accurately reconstructed from the output sequence. Notably, unlike standard machine translation, this constitutes an unbalanced sequence prediction problem, as the output sequences are typically much longer than the inputs.

\begin{figure}
\begin{center}
\includegraphics[width=\columnwidth]{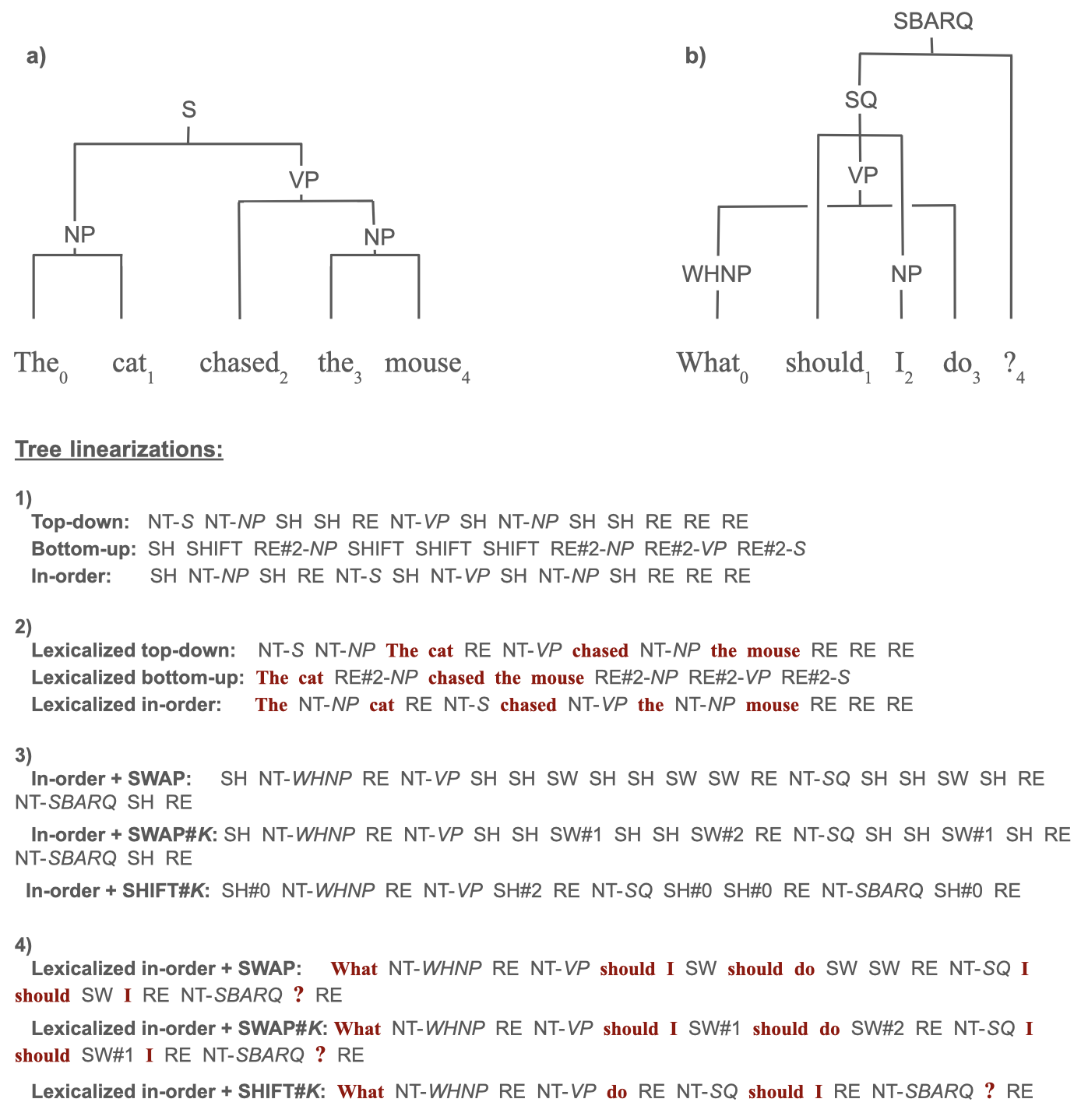}
\end{center}
\caption{Examples of continuous (a) and discontinuous (b) constituent trees, and different linearization strategies (1-4) for encoding those trees. SH = \textsc{Shift}, SH\#k = \textsc{Shift\#k}, NT$-{X}$ = \textsc{Non-Terminal-X}, RE = \textsc{Reduce}, RE\#k$-{X}$ = \textsc{Reduce\#k-X}, SW = \textsc{Swap}, and SW\#k = \textsc{Swap\#k}.}
\label{fig:linearizations}
\end{figure}

\paragraph{Constituent trees} A constituent tree $C$ consists of a sequence of words $w_0, \dots, w_{n-1}$ as its leaf nodes, with a hierarchy of constituents (or phrases) organized above them. Each constituent is represented as a tuple $(X, \mathcal{Y})$, where $X$ denotes the non-terminal label and $\mathcal{Y}$ indicates the set of word positions forming its yield or span. We define a constituent tree $C$ as \textit{continuous} if every constituent in the tree spans a contiguous subsequence of word positions (that is, a substring with no gaps). For example, the constituent tree in Figure~\ref{fig:linearizations}(a) illustrates such a continuous structure. In contrast, if any constituent in $C$ has a yield that includes non-consecutive word positions, the tree is considered \textit{discontinuous}. This is the case in Figure~\ref{fig:linearizations}(b), where the constituent $\text{(VP}, \{0, 3\})$ is disrupted by the words \textit{should}$_1$ and \textit{I}$_2$, which belong to the constituent $\text{(SQ}, \{1, 2\})$. As a result, the VP yield forms a sequence of non-adjacent positions, characterizing the structure as discontinuous.

\paragraph{Tree linearizations} Various strategies have been proposed to linearize constituent trees into sequences of tokens. Our approach builds upon the most successful paradigm, which employs transition-based algorithms to convert phrase-structure trees into sequences of actions. These actions, when applied according to a specific transition system, reconstruct the original tree structure. Among these methods, three transition-based algorithms have been particularly prominent in recent work on sequence-to-sequence constituent parsing~\citep{fernandez-gonzalez-gomez-rodriguez-2020-enriched,FERNANDEZGONZALEZ202343}: the \textit{top-down}~\citep{Dyer2016}, the \textit{bottom-up}~\citep{nonbinary}, and the \textit{in-order} transition systems~\citep{Liu2017}.

\paragraph{Transition systems} Transition systems are state machines that progress through a sequence of states via a set of transitions (or actions), ultimately arriving at a terminal state from which the output tree can be reconstructed. In the mentioned transition systems, each state is represented by a \textit{buffer}, initially containing all input words, and a \textit{stack}, which is initially empty and stores constituents and intermediate elements during parsing. Each system defines its own set of transitions for constructing constituent trees.

Specifically, the top-down transition system employs the following actions:
\begin{itemize}
    \item \textsc{Non-Terminal-X}: pushes a non-terminal symbol 
    \texttt{X} onto the stack,
    \item \textsc{Shift}: moves a word from the buffer onto the stack,
    \item \textsc{Reduce}: pops elements from the stack until a non-terminal node is found, then groups these elements into a new constituent placed on top of the stack.
\end{itemize}

In the bottom-up transition system, the available actions are:
\begin{itemize}
    \item \textsc{Shift}: transfers a word from the buffer to the stack,
    \item \textsc{Reduce\#k-X}: a parameterized action that removes $k$ elements from the stack and combines them into a constituent with the non-terminal label \texttt{X}, which is then pushed onto the stack.
\end{itemize}

Lastly, the in-order transition system utilizes actions similar to those in the top-down approach but with different application rules:
\begin{itemize}
    \item \textsc{Shift}: moves a word from the buffer to the stack,
    \item \textsc{Non-Terminal-X}: pushes a non-terminal label \texttt{X} onto the stack, but unlike in the top-down system, it is only valid if the first child of the upcoming constituent is already at the top of the stack,
    \item \textsc{Reduce}: pops items from the stack up to the first non-terminal node, which is also popped, and builds a new constituent from the collected elements on top of the stack.
\end{itemize}

\paragraph{Discontinuous transition systems} The transition systems described above are inherently limited to processing continuous constituent trees. To handle discontinuous structures, additional transitions are required. Following the approach of \citet{FERNANDEZGONZALEZ202343}, we extend continuous transition systems by incorporating actions capable of handling discontinuities. Specifically, we consider the alternative that incorporates a \textsc{Swap} action. This transition reorders the input sequence by moving the second element from the top of the stack back to the buffer. Such reordering enables the generation of any discontinuous tree using the standard set of continuous transitions, as any discontinuous constituent tree can be transformed into a continuous one by permuting word order. For example, the discontinuous structure in Figure~\ref{fig:linearizations}(b) can be converted into a continuous tree by repositioning the words \textit{should}$_1$ and \textit{I}$_2$ after the word \textit{do}$_3$. Figure~\ref{fig:example} demonstrates how the in-order transition system, extended with the \textsc{Swap} action, builds the discontinuous tree in Figure~\ref{fig:linearizations}(b), using buffer and stack manipulations as done in regular transition-based parsers. 

This transition system enhanced with the \textsc{Swap} action can be further extended by using a parameterized \textsc{Swap\#$k$} transition instead. This operation
%, also considered by~\citet{FERNANDEZGONZALEZ202343},
%in sequence-to-sequence constituent parsing, 
is equivalent to performing $k$ consecutive \textsc{Swap} actions (e.g., \textsc{Swap\#1} corresponds to a single \textsc{Swap}). The primary motivation for incorporating the \textsc{Swap\#$k$} transition rather than the standard \textsc{Swap} is to shorten the resulting transition sequences, as longer sequences tend to degrade performance on final tokens (a behavior observed in both transition-based parsers and sequence-to-sequence models).

\begin{figure}
\begin{center}
%\small
%\vspace*{13pt}
\begin{tabular}{@{\hskip 0.1pt}l@{\hskip 0.1pt}c@{\hskip 0.1pt}c@{\hskip 0.1pt}}
\hline
Transition & Stack & Buffer \\
\hline
\vspace*{3pt}
 & [ ] & [ What$_0$, should$_1$, I$_2$, do$_3$ , \textbf{?$_4$} ]\\
\vspace*{3pt}
\textsc{Shift} & [ What$_0$ ]& [ should$_1$, I$_2$, do$_3$ , \textbf{?$_4$} ]\\
\vspace*{3pt}
\textsc{NT-WHNP}& [ What$_0$, \textsc{WHNP} ] & [ should$_1$, I$_2$, do$_3$ , \textbf{?$_4$} ] \\
\vspace*{3pt}
\textsc{Reduce}& [ \textsc{WHNP}$_{(What_0)}$ ] & [ should$_1$, I$_2$, do$_3$ , \textbf{?$_4$} ] \\
\vspace*{3pt}
\textsc{NT-VP}& [ \textsc{WHNP}$_{(What_0)}$, \textsc{VP} ] & [ should$_1$, I$_2$, do$_3$ , \textbf{?$_4$} ] \\
\vspace*{3pt}
\textsc{Shift}& [ \textsc{WHNP}$_{(What_0)}$, \textsc{VP}, should$_1$ ] & [ I$_2$, do$_3$ , \textbf{?$_4$} ] \\
\vspace*{3pt}
\textsc{Shift}& [ \textsc{WHNP}$_{(What_0)}$, \textsc{VP}, should$_1$, I$_2$ ] & [ do$_3$ , \textbf{?$_4$} ] \\
\vspace*{3pt}
\textsc{Swap}& [ \textsc{WHNP}$_{(What_0)}$, \textsc{VP}, I$_2$ ] & [ should$_1$, do$_3$ , \textbf{?$_4$} ] \\
\vspace*{3pt}
\textsc{Shift}& [ \textsc{WHNP}$_{(What_0)}$, \textsc{VP}, I$_2$, should$_1$ ] & [ do$_3$ , \textbf{?$_4$} ] \\
\vspace*{3pt}
\textsc{Shift}& [ \textsc{WHNP}$_{(What_0)}$, \textsc{VP}, I$_2$, should$_1$, do$_3$ ] & [ \textbf{?$_4$}  ] \\
\vspace*{3pt}
\textsc{Swap}& [ \textsc{WHNP}$_{(What_0)}$, \textsc{VP}, I$_2$,  do$_3$ ] & [ should$_1$, \textbf{?$_4$}  ] \\
\vspace*{3pt}
\textsc{Swap}& [ \textsc{WHNP}$_{(What_0)}$, \textsc{VP}, do$_3$ ] & [ I$_2$, should$_1$, \textbf{?$_4$}  ] \\
\vspace*{3pt}
\textsc{Reduce}& [ \textsc{VP}$_{(WHNP,\ do_3)}$ ] & [ I$_2$, should$_1$, \textbf{?$_4$}  ] \\
\vspace*{3pt}
\textsc{NT-SQ}& [ \textsc{VP}$_{(WHNP,\ do_3)}$, \textsc{SQ} ] & [ I$_2$, should$_1$, \textbf{?$_4$}  ] \\
\vspace*{3pt}
\textsc{Shift}& [ \textsc{VP}$_{(WHNP,\ do_3)}$, \textsc{SQ}, I$_2$ ] & [ should$_1$, \textbf{?$_4$}  ] \\
\vspace*{3pt}
\textsc{Shift}& [ \textsc{VP}$_{(WHNP,\ do_3)}$, \textsc{SQ}, I$_2$, should$_1$ ] & [ \textbf{?$_4$}  ] \\
\vspace*{3pt}
\textsc{Swap}& [ \textsc{VP}$_{(WHNP,\ do_3)}$, \textsc{SQ}, should$_1$ ] & [ I$_2$, \textbf{?$_4$}  ] \\
\vspace*{3pt}
\textsc{Shift}& [ \textsc{VP}$_{(WHNP,\ do_3)}$, \textsc{SQ}, should$_1$, I$_2$ ] & [ \textbf{?$_4$}  ] \\
\vspace*{3pt}
\textsc{Reduce}& [ \textsc{SQ}$_{(VP,\ should_1,\ I_2)}$ ] & [ \textbf{?$_4$}  ] \\
\textsc{NT-SBARQ}& [ \textsc{SQ}$_{(VP,\ should_1,\ I_2)}$, \textsc{SBARQ} ] & [ \textbf{?$_4$}  ] \\
\vspace*{3pt}
\textsc{Shift}& [ \textsc{SQ}$_{(VP,\ should_1,\ I_2)}$, \textsc{SBARQ}, \textbf{?$_4$} ] & [ ] \\
\vspace*{3pt}
\textsc{Reduce}& [ \textsc{SBARQ}$_{(SQ,\ ?_4)}$ ] & [ ] \\
\hline
\end{tabular}
\caption{Transition sequence used by a transition-based parser to construct the discontinuous tree shown in Figure~\ref{fig:linearizations}(b), employing the in-order + \textsc{Swap} transition system along with the buffer and stack data structures. NT$-{X}$ = \textsc{Non-Terminal-X}.} \label{fig:example}   
\end{center}
\end{figure}

%Following \citet{FERNANDEZGONZALEZ202343}, 
Alternatively, we also explore a discontinuous transition system that introduces a \textsc{Shift\#$k$} transition instead of relying on \textsc{Swap} actions. This operation moves the $k$th word in the buffer to the stack, where \textsc{Shift\#0} corresponds to the standard \textsc{Shift} action. Figure~\ref{fig:example2} illustrates how the \textsc{Shift\#k} transition operates within the in-order transition system during a shift-reduce parsing process using a buffer and a stack. While incorporating the \textsc{Swap\#k} transition only slightly reduces sequence length, extending a transition system with \textsc{Shift\#k} yields transition sequences as short as those produced for continuous trees.

\begin{figure}
\begin{center}
%\small
%\vspace*{13pt}
\begin{tabular}{@{\hskip 0.1pt}l@{\hskip 0.1pt}c@{\hskip 0.1pt}c@{\hskip 0.1pt}}
\hline
Transition & Stack & Buffer \\
\hline
\vspace*{3pt}
 & [ ] & [ What$_0$, should$_1$, I$_2$, do$_3$ , \textbf{?$_4$} ]\\
\vspace*{3pt}
\textsc{Shift\#0} & [ What$_0$ ]& [ should$_1$, I$_2$, do$_3$ , \textbf{?$_4$} ]\\
\vspace*{3pt}
\textsc{NT-WHNP}& [ What$_0$, \textsc{WHNP} ] & [ should$_1$, I$_2$, do$_3$ , \textbf{?$_4$} ] \\
\vspace*{3pt}
\textsc{Reduce}& [ \textsc{WHNP}$_{(What_0)}$ ] & [ should$_1$, I$_2$, do$_3$ , \textbf{?$_4$} ] \\
\vspace*{3pt}
\textsc{NT-VP}& [ \textsc{WHNP}$_{(What_0)}$, \textsc{VP} ] & [ should$_1$, I$_2$, do$_3$ , \textbf{?$_4$} ] \\
\vspace*{3pt}
\textsc{Shift\#2}& [ \textsc{WHNP}$_{(What_0)}$, \textsc{VP}, do$_3$ ] & [ should$_1$, I$_2$ , \textbf{?$_4$} ] \\
\vspace*{3pt}
\textsc{Reduce}& [ \textsc{VP}$_{(WHNP,\ do_3)}$ ] & [ should$_1$, I$_2$, \textbf{?$_4$}  ] \\
\vspace*{3pt}
\textsc{NT-SQ}& [ \textsc{VP}$_{(WHNP,\ do_3)}$, \textsc{SQ} ] & [ should$_1$, I$_2$, \textbf{?$_4$}  ] \\
\vspace*{3pt}
\textsc{Shift\#0}& [ \textsc{VP}$_{(WHNP,\ do_3)}$, \textsc{SQ}, should$_1$ ] & [ I$_2$, \textbf{?$_4$}  ] \\
\vspace*{3pt}
\textsc{Shift\#0}& [ \textsc{VP}$_{(WHNP,\ do_3)}$, \textsc{SQ}, should$_1$, I$_2$ ] & [ \textbf{?$_4$}  ] \\
\vspace*{3pt}
\textsc{Reduce}& [ \textsc{SQ}$_{(VP,\ should_1,\ I_2)}$ ] & [ \textbf{?$_4$}  ] \\
\textsc{NT-SBARQ}& [ \textsc{SQ}$_{(VP,\ should_1,\ I_2)}$, \textsc{SBARQ} ] & [ \textbf{?$_4$}  ] \\
\vspace*{3pt}
\textsc{Shift\#0}& [ \textsc{SQ}$_{(VP,\ should_1,\ I_2)}$, \textsc{SBARQ}, \textbf{?$_4$} ] & [ ] \\
\vspace*{3pt}
\textsc{Reduce}& [ \textsc{SBARQ}$_{(SQ,\ ?_4)}$ ] & [ ] \\
\hline
\end{tabular}
\caption{Transition sequence used by a transition-based parser to construct the discontinuous tree shown in Figure~\ref{fig:linearizations}(b), employing the in-order + \textsc{Shift\#k} transition system along with the buffer and stack data structures. NT$-{X}$ = \textsc{Non-Terminal-X}.} \label{fig:example2}   
\end{center}
\end{figure}

\paragraph{Transition-based linearizations} The described transition systems\footnote{It is worth noting that the bottom-up and in-order transition systems (when implemented in transition-based parsers that explicitly define a buffer and a stack) require an additional flag along with a \textsc{Finish} transition to signal the end of the parsing process. However, in sequence-to-sequence models, which are entirely agnostic to such underlying data structures and do not construct subtrees during parsing, this flag and transition can be safely omitted.} were used by
previous studies~\citep{fernandez-gonzalez-gomez-rodriguez-2020-enriched,FERNANDEZGONZALEZ202343} for linearizing a constituent tree $C$, representing the syntactic structure of an input sentence $\mathbf{w}$, as a sequence of actions $\mathbf{y}$.
% = y_1, \dots, y_m$. 
For instance, Figure~\ref{fig:linearizations}(1) illustrates the token sequences derived from the continuous tree in Figure~\ref{fig:linearizations}(a) under each transition system. Similarly, Figure~\ref{fig:linearizations}(3) displays the sequences obtained from linearizing the discontinuous tree in Figure~\ref{fig:linearizations}(b) using transition-based strategies enhanced with \textsc{Swap}, \textsc{Swap\#$k$} and \textsc{Shift\#$k$} transitions. Please note that unlike task-specific transition-based parsers (which rely on a buffer, a stack, and a set of preconditions that must be satisfied with respect to these data structures), sequence-to-sequence models are entirely agnostic to such structures. Consequently, they generate sequence of tokens without considering the content of any data structure or enforcing any precondition.

\paragraph{Lexicalized linearizations} In order to better leverage pre-trained decoders, we further adapt the top-down, bottom-up and in-order linearization strategies by designing \textit{lexicalized} variants. Specifically, since input words are typically represented by \textsc{Shift} transitions in the output sequence, we further refine the action set by lexicalizing \textsc{Shift} transitions: i.e., each \textsc{Shift} is exclusively annotated with the specific word it transfers from the buffer to the stack. This makes each \textsc{Shift} and \textsc{Shift\#k} transitions explicitly represent a word from the input sentence. The resulting effect is illustrated in the sequences in Figures~\ref{fig:linearizations}(2) and (4). This lexicalization is particularly well suited for linearizations based on the transition systems employing the \textsc{Shift\#k} operation, where words are selected from the buffer by explicitly targeting a specific token. As an example, instead of applying a \textsc{Shift\#2} action to retrieve the word \textit{do} from the third position of the buffer, we simply use a \textsc{do} transition to perform the same operation. Although this variant makes it possible to linearize discontinuous trees using the same output sequence length and vocabulary as the corresponding continuous transition system, it introduces an inherent degree of lossiness. While words corresponding to regular \textsc{Shift} transitions are always mapped to the token ``Shift'' (regardless of whether they refer to the first word in the buffer), those encoding a \textsc{Shift\#k} action must unambiguously identify the targeted word in the buffer. However, this reference is not always accurate, since the same word may appear multiple times within a sentence, and in our implementation, the pointer always resolves to the first occurrence in the buffer. Consequently, the encoding variants that rely on a lexicalized \textsc{Shift\#k} transition are inherently lossy, achieving maximum F-scores of 98.95 on DPTB, 98.61 on NEGRA, and 98.94 on TIGER development sets.

Lexicalized \textsc{Shift} transitions were previously explored by~\citet{suzuki-etal-2018-empirical} and~\citet{fernandez-gonzalez-gomez-rodriguez-2020-enriched}. However, due to the large increase in output vocabulary (which negatively impacted early sequence-to-sequence models), these works only lexicalized \textsc{Shift} transitions for punctuation (specifically, commas and periods), yielding ``\textsc{Shift,}'' and ``\textsc{Shift.}'' transitions. Modern pre-trained encoder-decoder models
%, such as BART~\citep{lewis-etal-2020-bart} and T5~\citep{2020t5}, 
are naturally suited to this setting, as they are pre-trained to generate output sequences that share the same vocabulary as the input.

%can effectively manage the increased vocabulary. These models are trained by corrupting input text (e.g., by masking or deleting spans) and learning to reconstruct the original text, making them naturally suited to generate output sequences with the same vocabulary. 

\paragraph{Neural architecture} Pre-trained encoder–decoder Transformers, such as BART \citep{lewis-etal-2020-bart} and T5 \citep{2020t5}, follow the standard sequence-to-sequence architecture composed of a bidirectional encoder and an autoregressive decoder. The encoder processes the input sequence and produces contextualized representations for each token through stacked self-attention layers. The decoder then generates an output sequence token by token, attending both to previously generated tokens (via masked self-attention) and to the encoder representations (via cross-attention). These models are pre-trained on large text corpora using objectives designed to learn general sequence-to-sequence mappings (for example, denoising autoencoding in BART and text-to-text transformation tasks in T5) so that they can later be fine-tuned for a wide range of downstream tasks. In this work, we preserve the original neural architecture and adapt the models to the constituency parsing task through fine-tuning.

\paragraph{Parser implementation} 
Therefore, implementing a constituent parser for continuous trees with pre-trained encoder–decoder models proceeds as follows: the output vocabulary is extended with the non-\textsc{Shift} actions of each transition system, and the model is fine-tuned to generate an output sequence consisting of the input sentence with these actions inserted at the appropriate positions. For discontinuous trees, the procedure is analogous; however, linearizations based on \textsc{Swap} or \textsc{Swap\#$k$} transitions may be less appropriate for pre-trained encoder–decoder models, since words from the original sentence may be repeated multiple times in the output sequence (Figure~\ref{fig:linearizations}(4) ilustrates as the word ``should'' is repeated three times in the in-order+\textsc{Swap} and in-order+\textsc{Swap\#$k$} linearizations). In contrast, encodings that rely on the \textsc{Shift\#k} token should pose a task similar to that of the continuous setting, but with the added complexity of generating input words in a potentially different order.

\section{Experiments}
\label{sec:experiments}
\subsection{Setup} 
\paragraph{Data} To thoroughly evaluate our approach, we consider both continuous and discontinuous constituent treebanks. Specifically, we use the continuous English Penn Treebank (PTB)~\citep{marcus93} and its discontinuous counterpart, the Discontinuous Penn Treebank (DPTB)~\citep{evang-kallmeyer-2011-plcfrs}, adopting the standard data splits: Sections 2--21 for training, Section 22 for development, and Section 23 for testing. In addition, we include two German treebanks characterized by a higher degree of discontinuity: NEGRA~\citep{Skut1997} and TIGER~\citep{brants02}. For these, we follow widely adopted splits as defined by~\citet{dubey2003} for NEGRA and~\citet{seddah-etal-2013-overview} for TIGER. In all experiments, Part-of-Speech (PoS) tag information is excluded.

\paragraph{Pre-trained encoder-decoder transformers}  
Our experiments incorporate both the \texttt{base} and \texttt{large} variants of BART~\citep{lewis-etal-2020-bart}, as well as the \texttt{large} version of T5 \citep{2020t5}, for English parsing tasks. For German, we employ the multilingual mBART model~\citep{liu-etal-2020-multilingual-denoising}, which is pre-trained using a denoising objective over multilingual corpora. The architecture and hyperparameters used to fine-tune pre-trained models for the constituent parsing task are detailed in Table~\ref{tab:hyperparameters}.

% \begin{table}[h]
% \begin{footnotesize}
% \centering
% \begin{tabular}{@{\hskip 0pt}lc@{\hskip 0pt}}
% \multicolumn{2}{c}{\textbf{Architecture and optimizer hyper-parameters}} \\
% Transformer Encoder layers & 6 \\ 
% Transformer Encoder size & 256 \\
% Transformer Decoder layers & 6 \\ 
% Transformer Decoder size & 256 \\
% Heads per self-attention layer & 4 \\ 
% RoBERTa embedding dimension & 1024\\
% GottBERT embedding dimension & 768\\
% Dropout & 0.3 \\
% %MLP layers & 1 \\
% %MLP activation function & ELU \\
% %Arc MLP size & 512 \\ 
% %Label MLP size & 128 \\
% %UNK replacement probability & 0.5 \\
% Beam size & 10 \\
% %\midrule
% %\textbf{Adam optimizer hyper-parameters} &\\
% %\midrule
% Optimizer & Adam \citep{Adam} \\
% Loss & cross-entropy \\
% $\beta_1$ & 0.9 \\
% $\beta_2$ & 0.98 \\
% %$\epsilon$ & 1$e^{-8}$\\
% Learning rate & 5$e^{-4}$ \\
% Learning rate scheduler & Inverse square root \\
% Warm-up initial learning rate & 1$e^{-7}$ \\
% Warm-up updates & 4000 \\
% Minimum learning rate & 1$e^{-9}$ \\
% Label smoothing & 0.01 \\
% Batch size & 3584 \\
% Training epochs & 80 \\
% %Decay rate & 0.75 \\
% %Gradient clipping & 5.0 \\
% \end{tabular}
% \setlength{\abovecaptionskip}{4pt}
% \caption{Models hyper-parameters.}
% \label{tab:hyperparameters}
% \end{footnotesize}
% \end{table}

\begin{table}[h]
	\begin{footnotesize}
		\centering
		\begin{tabular}{@{\hskip 0pt}lcccc@{\hskip 0pt}}
			%\multicolumn{5}{c}{\textbf{Architecture and optimizer hyper-parameters}} \\
			
			& \textbf{BART-base} & \textbf{BART-large} & \textbf{T5} & \textbf{mBART} \\
			
			Transformer Encoder layers & 6 & 12 & 24 & 12 \\ 
			Transformer Encoder size & 768 & 1024 & 1024 & 1024 \\
			Transformer Decoder layers & 6 & 12 & 24 & 12 \\ 
			Transformer Decoder size & 768 & 1024 & 1024 & 1024 \\
			Heads per self-attention layer & 12 & 16 & 16 & 16 \\ 
			FFN dimension & 3072 & 4096 & 4096 & 4096 \\
            Activation function & GELU & GELU & ReLU & ReLU \\
			Dropout & 0.1 & 0.1 & 0.1 &  0.1\\
			Beam size & 10 & 10 & 10 & 10 \\
			
			Optimizer & AdamW \citep{Adam} & AdamW  & AdamW & AdamW  \\
			Loss & cross-entropy & cross-entropy & cross-entropy & cross-entropy \\
			$\beta_1$ & 0.9 & 0.9 & 0.9 & 0.9 \\
			$\beta_2$ & 0.999 & 0.999 & 0.999 & 0.999 \\
		    $\epsilon$  & 1$e^{-8}$ & 1$e^{-8}$ & 1$e^{-8}$ & 1$e^{-8}$ \\
			Learning rate & 1$e^{-5}$ & 1$e^{-5}$ & 1$e^{-5}$ & 1$e^{-5}$ \\
			Weight decay & 0.01 & 0.01 & 0.01 & 0.01 \\
			Learning rate scheduler & Linear decay & Linear decay  & Linear decay & Linear decay \\
			Warm-up ratio & 0.1 & 0.1 & 0.1 & 0.1 \\
			Train Batch size & 32 & 16 & 4 & 8 \\
			Eval Batch size & 32 & 8 & 4 & 8 \\
			Training epochs & 100 & 100 & 100 & 120 \\

            Max sequence length & 1024 & 1024 & 1024 & 1024 \\
			Temperature & 0.7 & 0.7 & 0.7 & 0.7 \\
			
		\end{tabular}
		\setlength{\abovecaptionskip}{4pt}
		\caption{Architecture and optimizer hyper-parameters for each pre-trained encoder-decoder model.}
		\label{tab:hyperparameters}
	\end{footnotesize}
\end{table}

\paragraph{Evaluation} We adhere to standard evaluation protocols and report F-scores using the EVALB script\footnote{\url{https://nlp.cs.nyu.edu/evalb/}} for the continuous PTB, excluding punctuation. For discontinuous treebanks, we employ the DISCODOP toolkit\footnote{\url{https://github.com/andreasvc/disco-dop}}~\citep{Cranenburgh2016}, which similarly ignores punctuation and root symbols. In addition to overall F-score, DISCODOP provides a Discontinuous F-score (DF1), which evaluates performance specifically on discontinuous constituents. For all experiments, we report the average score and standard deviation across three runs with different random initializations.

%Finally, our approach was fully tested on an Intel(R) Core(TM) i9-10920X CPU @ 3.50GHz with a single 24 GB TESLA P40 GPU.

\subsection{Results}
\paragraph{Continuous treebank} Table~\ref{tab:con} reports accuracies on the development and test splits, together with a comparison against state-of-the-art continuous constituency parsers, including all existing sequence-to-sequence models. Our method attains competitive performance across all tree linearization schemes, and we observe no significant differences between top-down and in-order strategies. In line with the findings of \cite{FERNANDEZGONZALEZ202343}, bottom-up linearization also under this architecture yields inferior results compared to their top-down and in-order counterparts. These results suggest that excessively extending the original vocabulary with new symbols (as occurs in the bottom-up strategy) negatively affects final performance, as also observed in the encoder-only setting by \citet{FERNANDEZGONZALEZ202343}.

With respect to the pre-trained encoder–decoder backbone, BART$_\textsc{large}$ achieves the highest scores, outperforming the smaller BART$_\textsc{base}$ variant, as expected, as well as the large version of T5. Finally, it is worth noting that the large version of the T5 model is also surpassed by the BART$_\textsc{base}$ variant, and the bottom-up linearization applied to T5 results in considerably poorer performance.

Compared with other sequence-to-sequence constituency parsers, our method surpasses all existing models by a substantial margin. Moreover, when contrasted with the strongest task-specific approaches, the top-down and in-order tree linearizations achieve performance on par with models augmented with the pre-trained language model XLNet$_\textsc{large}$ \citep{XLNet}, and clearly outperform those based on BERT \citep{devlin-etal-2019-bert}.

%In fact, our approach is on par, for instance, with \cite{attachjuxtapose} (a purely transition-based parser) and \cite{tian-etal-2020-improving} (a chart-based model) when BERT$_\textsc{Large}$ is used instead.

\begin{table}[t]
\begin{center}
\centering
\begin{tabular}{@{\hskip 2pt}l@{\hskip -30pt}l@{\hskip 6pt}l@{\hskip 0pt}}
\textbf{Parsers}  &  &\textbf{PTB} \\
\hline
\textit{\footnotesize (task-specific approaches)} &  & \\
\citet{kitaev-etal-2019-multilingual} + BERT$_\textsc{Large}{}$ & & 95.59 \\
\citet{attachjuxtapose} + BERT$_\textsc{Large}$ & & 95.79 \\
\citet{zhou-zhao-2019-head} + dep + BERT$_\textsc{Large}{}$ & & 95.84 \\
\citet{tian-etal-2020-improving} + PoS + BERT$_\textsc{Large}$ & & 95.86 \\
\citet{zhou-zhao-2019-head} + dep + XLNet$_\textsc{large}$ & & 96.33 \\
\citet{attachjuxtapose} + XLNet$_\textsc{large}$ & & 96.34 \\
\citet{mrini-etal-2020-rethinking} + dep + PoS + XLNet$_\textsc{large}$ & & 96.38 \\
\citet{tian-etal-2020-improving} + XLNet$_\textsc{large}$ & & 96.36\\
\citet{tian-etal-2020-improving} + PoS + XLNet$_\textsc{large}$ & & \textbf{96.40}\\
%Fernández-G\&Gómez-R \citep{multipointer}+dep+ BERT$_\textsc{Large}{}$ & & 95.23\\ 
\hline
\textit{\footnotesize (sequence-to-sequence models)} & &  \\
\citet{Vinyals2015} &  & 88.3\hphantom{0}  \\
\citet{Vinyals2015} + ensemble &  & 90.5\hphantom{0}  \\
\citet{LiuS2S17} & & 90.5\hphantom{0}  \\
\citet{Ma2017} + ensemble & & 90.6\hphantom{0}  \\
\citet{Kamigaito2017} + ensemble & & 91.5\hphantom{0}  \\
F\&R \cite{fernandez-gonzalez-gomez-rodriguez-2020-enriched} & & 91.6\hphantom{0}  \\
\citet{Liu2018} + ensemble & & 92.3\hphantom{0}  \\
\citet{Suzuki2018} + ensemble + LM-rerank & & 94.32  \\
%\citet{Vaswani2017} & & 91.3\hphantom{0} \\
F\&R \cite{FERNANDEZGONZALEZ202343} + RoBERTa$_\textsc{Large}$ + in-order & & 95.84 \\
F\&R \cite{FERNANDEZGONZALEZ202343} + RoBERTa$_\textsc{Large}$ + top-down & & 95.78 \\
\textbf{This work:} \hspace{8.5cm} &\textbf{(dev)} &\textbf{(test)}  \\
\ \ \ \ \textbf{BART$_\textsc{base}$ + lexicalized top-down} & 95.16\tiny{$\pm$0.18} & 95.69\tiny{$\pm$0.02}\\
\ \ \ \ \textbf{BART$_\textsc{base}$ + lexicalized bottom-up} & 94.26\tiny{$\pm$0.09} & 94.52\tiny{$\pm$0.02}\\
\ \ \ \ \textbf{BART$_\textsc{base}$ + lexicalized in-order} & 95.36\tiny{$\pm$0.15} & 95.61\tiny{$\pm$0.04}\\
\ \ \ \ \textbf{BART$_\textsc{large}$ + lexicalized top-down} & 95.75\tiny{$\pm$0.05} & 96.23\tiny{$\pm$0.06}\\
\ \ \ \ \textbf{BART$_\textsc{large}$ + lexicalized bottom-up} & 95.09\tiny{$\pm$0.06} & 95.36\tiny{$\pm$0.11}\\
\ \ \ \ \textbf{BART$_\textsc{large}$ + lexicalized in-order} & \textbf{95.76}\tiny{$\pm$0.03} & \textbf{96.24}\tiny{$\pm$0.02}\\
\ \ \ \ \textbf{T5$_\textsc{large}$ + lexicalized top-down} & 95.50\tiny{$\pm$0.03} & 95.55\tiny{$\pm$0.03}\\
\ \ \ \ \textbf{T5$_\textsc{large}$ + lexicalized bottom-up} & 91.23\tiny{$\pm$0.02} & 90.83\tiny{$\pm$0.16}\\
\ \ \ \ \textbf{T5$_\textsc{large}$ + lexicalized in-order} & 95.43\tiny{$\pm$0.06} & 95.45\tiny{$\pm$0.12}\\
\end{tabular}
\centering
\setlength{\abovecaptionskip}{4pt}
\caption{F-score comparison of state-of-the-art constituent parsers on the PTB test set, without the use of gold or predicted PoS tags. The second section of the table is dedicated exclusively to sequence-to-sequence models. Additionally, we report parser performance on the PTB development set for our models. The abbreviation F\&R \cite{FERNANDEZGONZALEZ202343} refers to work by Fernández-González and Gómez-Rodríguez \cite{FERNANDEZGONZALEZ202343}, and we report the performance obtained by both non-lexicalized top-down and in-order linearizations. Parsers leveraging additional dependency information are marked with \textit{+dep}; those that combine multiple trained models through ensembling are denoted \textit{+ensemble} and models using language models to rerank predicted trees are labeled \textit{+LM-rerank}. Models using predicted PoS tags as auxiliary input are annotated with \textit{+PoS}. 
Finally, systems that initialize the encoder with pre-trained language models such as BERT$_\textsc{Large}$~\citep{devlin-etal-2019-bert}, RoBERTa$_\textsc{Large}$ \citep{roberta} or XLNet$_\textsc{large}$~\citep{XLNet} are labeled \textit{+BERT$_\textsc{Large}$/+RoBERTa$_\textsc{Large}$/+XLNet$_\textsc{large}$}. }
\label{tab:con}
\end{center}
\end{table}

\paragraph{Discontinuous treebanks} We further evaluate the proposed sequence-to-sequence model with top-down+\textsc{Swap}, in-order+\textsc{Swap}, in-order+\textsc{Swap\#k} and in-order+\textsc{Shift\#k}  linearizations on the test splits of discontinuous treebanks (Table~\ref{tab:disc}). For comparison, we include the strongest task-specific methods reported to date, as well as the only existing sequence-to-sequence model for discontinuous constituency parsing \cite{FERNANDEZGONZALEZ202343}, enabling a comprehensive assessment against the current state of the art. For the work by \citet{FERNANDEZGONZALEZ202343}, we report the different results obtained using the equivalent non-lexicalized linearizations.

Consistent with the results on the continuous benchmark, the in-order linearization (augmented with the \textsc{Swap} transition) achieves higher overall and discontinuous F-scores than the top-down variant across all three datasets, with particularly notable improvements on the German treebanks. Among the in-order variants combined with different strategies for constructing discontinuities, 
%(namely by introducing the \textsc{Swap}, \textsc{Swap\#K}, and \textsc{Shift\#K} actions), 
we observe that the addition of the \textsc{Shift\#K} transition yields the best overall scores on the German treebanks and performs on par with the regular \textsc{Swap} on the DPTB. In contrast, the variant using the \textsc{Swap\#K} token is substantially less accurate, likely because it requires extending the original vocabulary with significantly more symbols than the other variants.
However, when considering the discontinuous F-score, the strategy based on the \textsc{Swap} transition achieves the best results on TIGER and DPTB, while the \textsc{Shift\#K}-based variant only stands out in constructing discontinuities in NEGRA and performs notably poorly on DPTB. Therefore, although the in-order \textsc{Shift\#K}-based variant appears to be more suitable for a pre-trained encoder–decoder model, the lossy nature of this linearization (arising from the possibility of repeated words in the buffer) limits its ability to achieve the same performance in constructing discontinuous constituents as in the continuous case.

With respect to the other existing sequence-to-sequence model, our approach generally achieves lower results, particularly on the German datasets. This can be explained by the fact that \citet{FERNANDEZGONZALEZ202343} incorporate structural information into the Transformer architecture (whereas our approach relies on the original architecture without task-specific enhancements) and, for the German treebanks, employ the language-specific pre-trained model GottBERT$_\textsc{Base}$ (while we use a multilingual pretrained encoder-decoder model). Regarding the discontinuous linearizations, a different trend can be observed. In their work, the \textsc{Shift\#k}-based variant yields the worst results, with a substantial gap compared to the other approaches, whereas in our experiments it achieves the best scores (since in our framework  it can be implemented exactly as in the continuous in-order setting). Conversely, the \textsc{Swap\#k}-based strategy performs worst in our approach, while in their work it is the best option on DPTB and the second-best on the German treebanks. 

%A ESTUDIA EN EL ANALISIS
%This difference can largely be attributed to the structural information injected by their novel Transformer architecture, whereas our approach relies on the original architecture without task-specific enhancements. In contrast to approaches based on encoder-only models, the in-order+\textsc{Shift#K} variant in our framework can be implemented exactly as in the continuous in-order setting, leading to strong performance. However, the less natural \textsc{Swap}- and \textsc{Swap#K}-based alternatives require structural information that is not available in our approach.

Compared to task-specific discontinuous parsers, our \textsc{Swap}-based strategy improves over the best specialized parser on DPTB \cite{fernandezgonzalez2021reducing}. However, on the more challenging German treebanks, they only surpass approaches that do not incorporate language model enhancements.

%Regarding the accuracy on discontinuities, the top-down linearization obtains the best F-scores.

%With respect to the alternative with the \textsc{Swap\#k} token for shortening output sequences, we notice that it suffers a significant drop in both overall and disonctinuous F-score with respect to using the regular \textsc{Swap} action. On the other hand, the variant with the \textsc{Shift\#k} transition achieves the best overall F-scores on German treebanks and it is on par to the use of the \textsc{Swap} token on the DPTB. Regarding the performance on discontinuities, the variant with the \textsc{Shift\#k} underperforms the linearizations with the \textsc{Swap} action on TIGER and DPTB.

%Regarding other sequence-to-sequence models, our approach is outperformed by the system by \cite{FERNANDEZGONZALEZ202343}, which leverages specific pre-trained encoder-only models for German and English. Specially for German, the use of a specific pre-traiend models yields a notable impact in parsers performance in comparison to using a multilingual pre-trained model as it is our case. Regarding task specific models, 

Overall, our approach achieves the best reported F-score among sequence-to-sequence models on continuous treebanks and surpasses task-specific models enhanced with BERT. When discontinuities are present, it yields competitive performance when they are relatively sparse, outperforming all task-specific parsers on DPTB. Nevertheless, it struggles on German treebanks, where additional structural information is crucial for constructing discontinuous constituents.

\begin{table*}[h]
\small
\centering
\begin{tabular}{@{\hskip 2pt}l@{\hskip 4pt}c@{\hskip 3pt}cc@{\hskip 3pt}cc@{\hskip 3pt}c@{\hskip 2pt}}
& \multicolumn{2}{c}{\textbf{TIGER}}
& \multicolumn{2}{c}{\textbf{NEGRA}}
& \multicolumn{2}{c}{\textbf{DPTB}}
\\
\textbf{Parser} & \textbf{F1} & \textbf{DF1} & \textbf{F1} & \textbf{DF1} & \textbf{F1} & \textbf{DF1} \\
\hline
\footnotesize{\textit{(task-specific approaches)}} & & & & &  & \\
\citet{coavoux2019b} & 82.5 & 55.9  & 83.2 & 56.3 & 90.9 & 67.3  \\
\citet{coavoux2019a} & 82.7 & 55.9 & 83.2 & 54.6  & 91.0 & 71.3  \\
\citet{stanojevic-steedman-2020-span} & 83.4 & 53.5 & 83.6 & 50.7 & 90.5 & 67.1 \\
\citet{Corro2020SpanbasedDC} + BERT$_\textsc{X}$ & 90.0 & 62.1 & 91.6 & 66.1 & 94.8 & 68.9 \\
F\&R \cite{DiscoPointer} & 85.7 & 60.4 & 85.7 & 58.6 & - & -  \\
R\&M \cite{morbitz2020supertaggingbased} + BERT$_\textsc{Base}$ & 88.3 & 69.0 & 90.9 & 72.6 & 93.3 & 80.5 \\
F\&R \cite{multipointer} + dep + BERT$_\textsc{Base}$ & 89.8 & \textbf{71.0} & 91.0 & \textbf{76.6} & - & -  \\

F\&R  \cite{fernandezgonzalez2021reducing} + BERT$_\textsc{Large}$ & \textbf{90.5} & 68.1 & \textbf{92.0} & 67.9 & 94.7 & 72.9 \\
F\&R \cite{fernandezgonzalez2021reducing} +  XLNet$_\textsc{large}$ & - & - & - & - & \textbf{95.1} & \textbf{74.1}  \\
%\citet{ruprecht-2022-improving} + BERT$_\textsc{Large}$ & \textbf{91.6} & \textbf{75.4} & \textbf{93.9} & \textbf{79.1} & 94.9 & \textbf{82.4} \\
\hline
\footnotesize{\textit{(sequence-to-sequence models)}} & & & & &  & \\
%F\&R \cite{FERNANDEZGONZALEZ202343} + RoBERTa$_\textsc{Large}$& - & - & - & - & \textbf{95.47} & \textbf{83.80} \\
%F\&R \cite{FERNANDEZGONZALEZ202343} + GottBERT$_\textsc{Base}$& \textbf{88.53} & \textbf{67.76} & \textbf{89.08} & \textbf{67.06} & - & - \\
F\&R \cite{FERNANDEZGONZALEZ202343} + GottBERT$_\textsc{Base}$&  &  &  &  &  &  \\
\ \ \ \  \ \ \ \  \ \ \ \  \ \ \ \  \ \ \ \  /RoBERTa$_\textsc{Large}$&  &  &  &  &  &  \\
\ \ \ \ top-down + \textsc{Swap}   & 88.28 & \textbf{67.95} & 88.59 & \textbf{67.43} & 95.37 & \textbf{83.85}\\
\ \ \ \ in-order + \textsc{Swap} & \textbf{88.53} & 67.76 & \textbf{89.08} & 67.06 & 95.47 & 83.80 \\
\ \ \ \ in-order + \textsc{Swap\#k}  & 88.36 & 65.68 & 88.93 & 65.38 & \textbf{95.48} & 82.86 \\
\ \ \ \ in-order + \textsc{Shift\#k} & 87.10 & 54.27 & 86.76 & 46.86 & 94.96 & 69.17 \\
\textbf{This work:} & \tiny{$\pm$0.07} & \tiny{$\pm$0.32} & \tiny{$\pm$0.1} & \tiny{$\pm$0.57} & \tiny{$\pm$0.13} & \tiny{$\pm$0.74}\\
\ \ \ \ \textbf{lexicalized top-down} + \textsc{Swap}  & 86.61 & 59.18 & 85.34 & 53.14 & 95.06 & 74.39\\
& \tiny{$\pm$0.23} & \tiny{$\pm$0.57} & \tiny{$\pm$0.41} & \tiny{$\pm$0.96} & \tiny{$\pm$0.15} & \tiny{$\pm$1.78}\\
\ \ \ \ \textbf{lexicalized in-order} + \textsc{Swap} & 87.28 & \textit{61.95} & 86.75 & 57.71 & \textit{95.14} & \textit{75.14} \\
& \tiny{$\pm$0.07} & \tiny{$\pm$0.38} & \tiny{$\pm$0.38} & \tiny{$\pm$0.2} & \tiny{$\pm$0.07} & \tiny{$\pm$2.19}\\
\ \ \ \ \textbf{lexicalized in-order} + \textsc{Swap\#k} & 84.57 & 50.74 & 83.02 & 40.82 & 94.58 & 67.94 \\
& \tiny{$\pm$0.09} & \tiny{$\pm$0.42} & \tiny{$\pm$0.49} & \tiny{$\pm$0.09} & \tiny{$\pm$0.02} & \tiny{$\pm$0.28}\\
\ \ \ \ \textbf{lexicalized in-order} + \textsc{Shift\#k} & \textit{87.52} & 57.53 & \textit{88.54} & \textit{58.53} & 95.10 & 62.95 \\
\end{tabular}
\centering
\setlength{\abovecaptionskip}{4pt}
\caption{
F-score and Discontinuous F-score (DF1) comparison of leading discontinuous constituent parsers on the TIGER, NEGRA, and DPTB test sets, without the use of gold or predicted PoS tags. 
R\&M \cite{morbitz2020supertaggingbased} refers to the model by \citet{morbitz2020supertaggingbased}, and F\&R denotes a series of works by Fernández-González and Gómez-Rodríguez. In particular, for the sequence-to-sequence approach introduced by \citet{FERNANDEZGONZALEZ202343}, we examine non-lexicalized linearization variants.
%that incorporate additional dependency-based features are denoted with \textit{+dep}. Models initialized with pre-trained language representations—such as BERT$_\textsc{base}$, BERT$_\textsc{Large}$~\citep{devlin-etal-2019-bert}, RoBERTa$_\textsc{Large}$ \citep{roberta}, or XLNet$_\textsc{large}$~\citep{XLNet} are labeled with \textit{+BERT$_\textsc{base}$/+BERT$_\textsc{Large}$/+RoBERTa$_\textsc{Large}$/+XLNet$_\textsc{large}$} (
We adopt the same abbreviations as in Table~\ref{tab:con} to indicate additional enhancements of parsers. Additionally, a model initialized with the German pre-trained language model GottBERT$_\textsc{base}$~\citep{gottbert} is labeled as \textit{+GottBERT$_\textsc{base}$}, and the notation \textit{+BERT$_\textsc{X}$} is used when the specific size of the BERT model is unspecified.
%). 
 For English, only the BART$_\textsc{large}$ variant is employed in our models. Finally, we mark the best results for task-specific and sequence-to-sequence parsers in bold, while italic indicates the best results among our lexicalized linearizations.
}
\label{tab:disc}
\end{table*}

\subsection{Analysis}
\label{sec:analysis}

To gain further insight into the performance differences between linearizations and encoder–decoder models, we conduct an error analysis considering structural factors and sentence length on the development splits of the continuous PTB and the concatenation of the three discontinuous treebanks. We also include the outputs of the sequence-to-sequence model by \citet{FERNANDEZGONZALEZ202343} for comparison; while it introduces structural information into a pre-trained encoder-only architecture, it employs equivalent (but non-lexicalized) linearizations.

\begin{figure}
\begin{center}
\includegraphics[width=\columnwidth]{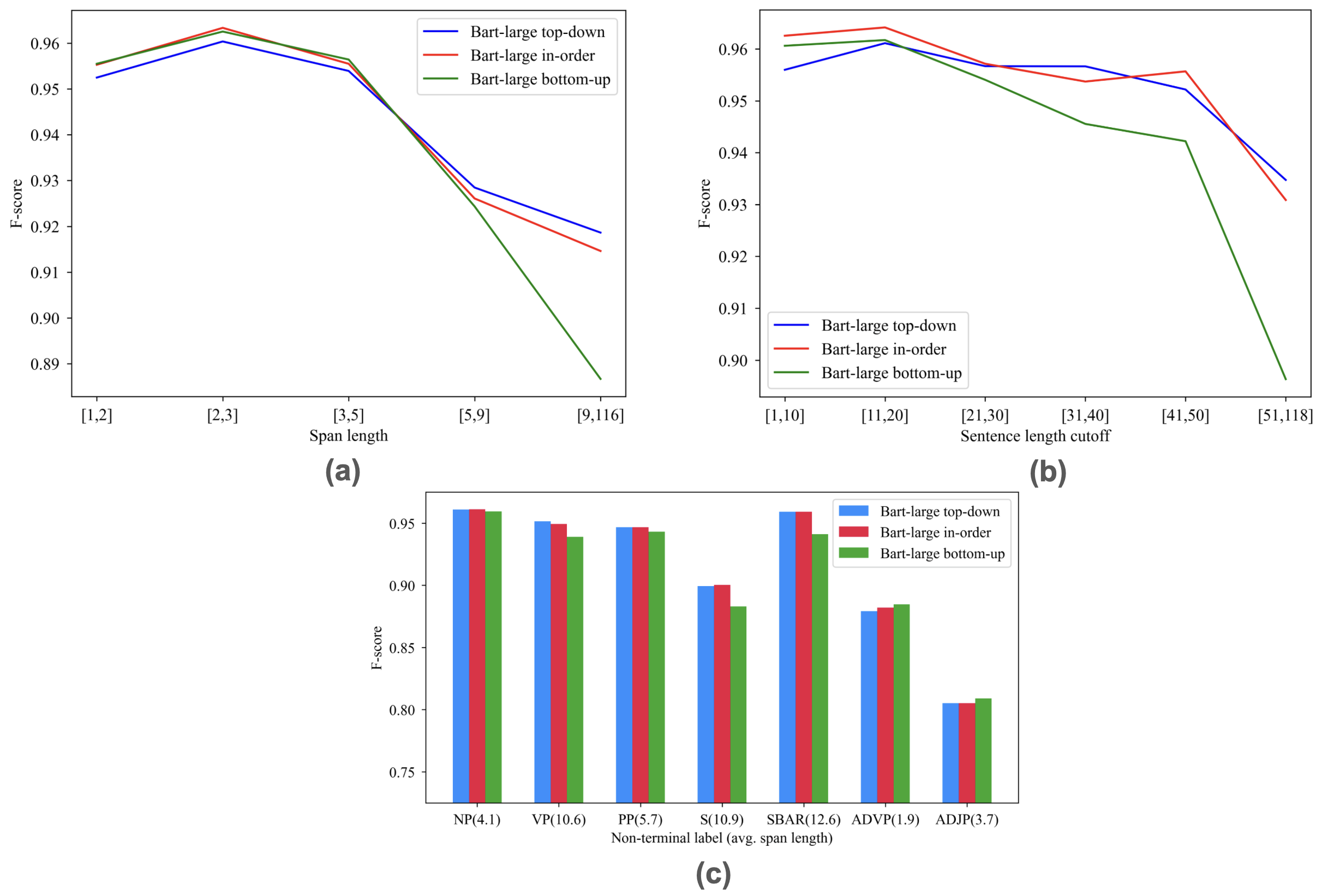}
\end{center}
\caption{F-score achieved by continuous linearizations on BART$_\textsc{Large}$ across structural factors and sentence lengths.}
\label{fig:analysis1}
\end{figure}

\paragraph{Linearization comparison in continuous parsing} For each proposed continuous tree linearization using a shared encoder--decoder model (BART$_\textsc{Large}$), we show  the F-score for span identification across span length intervals grouped by frequency in Figure~\ref{fig:analysis1}(a),  the performance across sentence length thresholds based on width in Figure~\ref{fig:analysis1}(b), and the F-score for the most frequent non-terminal labels (with average span length indicated in brackets) in Figure~\ref{fig:analysis1}(c). From these results, we can conclude that:
\begin{itemize}
    \item Error propagation is visible in Figures~\ref{fig:analysis1}(a) and (b) for sequence-to-sequence models. The sequential prediction process may cause early mistakes to influence later decisions, leading to additional errors in the encoded constituent tree.

    \item The in-order and bottom-up strategies outperform the top-down approach on spans shorter than 5. However, they fall behind on larger spans, with the bottom-up method showing a particularly sharp decline in F-score for spans longer than 9. This degradation is mainly due to the need to predict actions of the form \textsc{Reduce\#k-X} with larger values of $k$, which are less frequent in the training data and therefore harder to learn.

    \item Although it is not the best-performing method on shorter spans, the top-down approach maintains stronger performance on longer spans and exhibits the smallest overall degradation.

    \item The bottom-up variant performs well on shorter sentences, even surpassing the top-down approach in this setting. However, its performance declines significantly as sentence length increases, likely because longer sentences involve larger constituents, which are more challenging for this strategy.

    \item The in-order variant outperforms the top-down approach across most sentence lengths, except for sentences in the 31-40 range and the longest ones (longer than 51 tokens).

    \item While the bottom-up approach outperforms the other variants in identifying the most frequent short constituents (e.g., ADVP and ADJP), it shows substantial drops in accuracy for constituents such as VP, S, and SBAR. This can be attributed to the fact that these structures typically span longer sequences, making them harder to model with this linearization.

    \item Finally, the in-order and top-down variants exhibit largely comparable performance across constituent types.
\end{itemize}

\begin{figure}
\begin{center}
\includegraphics[width=\columnwidth]{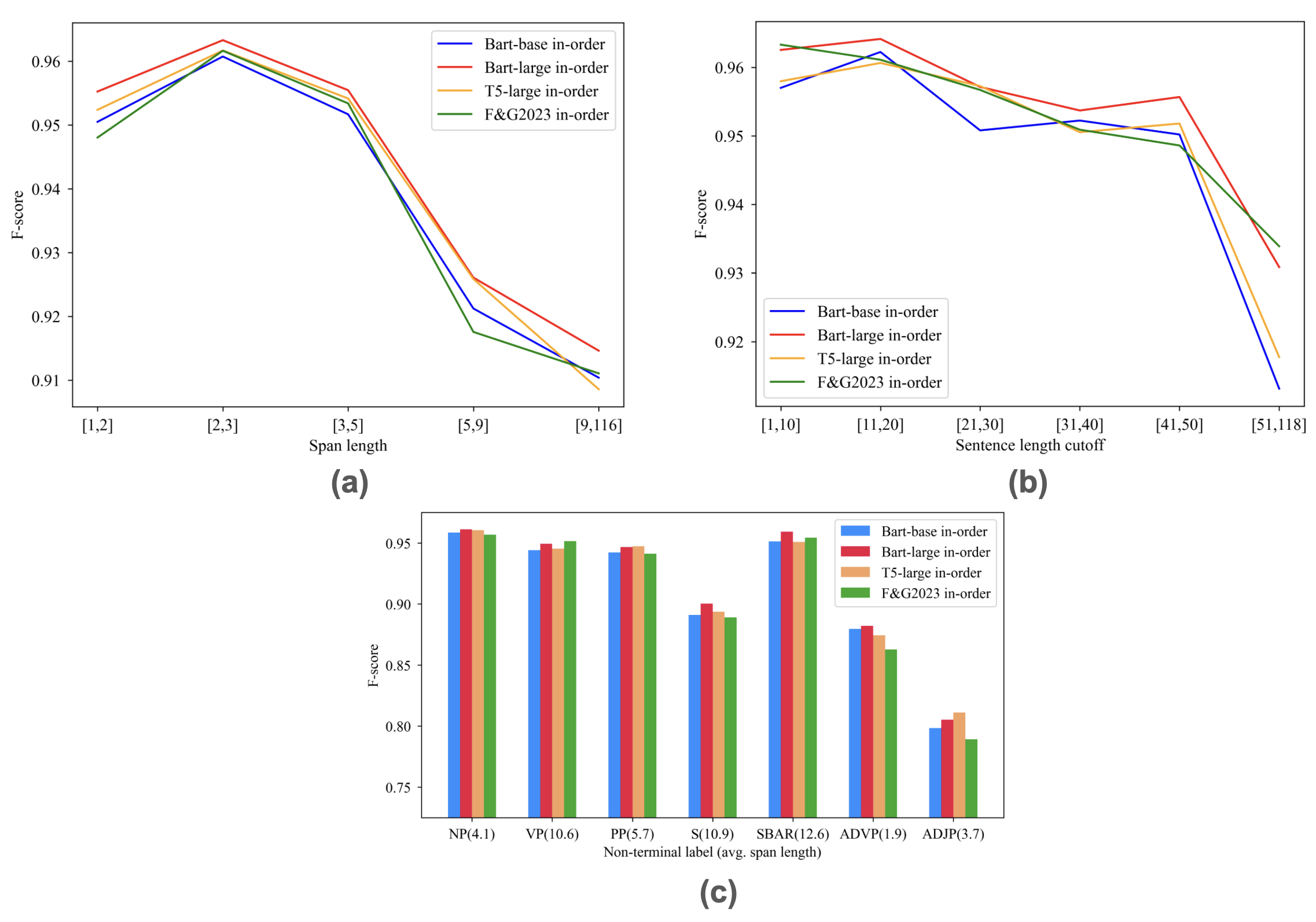}
\end{center}
\caption{F-score of  the lexicalized in-order linearization across pre-trained encoder–decoder models (including the encoder-only model  with the non-lexicalized in-order strategy by \citet{FERNANDEZGONZALEZ202343} as ``F\&G2023'') with respect to structural factors and sentence length.}
\label{fig:analysis2}
\end{figure}

\paragraph{Model comparison in continuous parsing} 
We further investigate the impact of different neural architectures under a fixed tree linearization. To this end, Figure~\ref{fig:analysis2}(a) presents the F-score for span identification across span lengths, Figure~\ref{fig:analysis2}(b) shows performance across sentence length thresholds, and Figure~\ref{fig:analysis2}(c) reports the F-score for the most frequent non-terminal labels, all using the in-order linearization across different models. Specifically, the comparison includes the pre-trained encoder--decoder models BART$_\textsc{Base}$, BART$_\textsc{Large}$, and T5$_\textsc{Large}$, as well as the encoder-only RoBERTa$_\textsc{Large}$ model (with a non-lexicalized in-order linearization) from \citet{FERNANDEZGONZALEZ202343} (referred to as ``F\&G2023''). Based on these results, we can draw the following conclusions:
\begin{itemize}
    \item Overall, BART$_\textsc{Large}$ achieves the best performance across all span and sentence length ranges.

    \item BART$_\textsc{Large}$ is only surpassed by F\&G2023 on very short sentences (fewer than 10 tokens) and the longest ones (over 51 tokens). The latter likely reflects the importance of additional structural information, as incorporated in F\&G2023, for modeling long sequences.

    \item T5$_\textsc{Large}$ generally ranks as the second-best model across most span length intervals, except for the largest constituents, where it performs the worst. In terms of sentence length, it is only the second-best model for sentences between 41 and 50 tokens, and is even outperformed by BART$_\textsc{Base}$ for lengths between 11--20 and 31--40. These inconsistencies lower its overall performance relative to BART$_\textsc{Large}$, despite their architectural similarities.

    \item Interestingly, despite being the smallest model, BART$_\textsc{Base}$ outperforms F\&G2023 on the shortest constituents and those with span lengths between 5 and 9, and also surpasses both F\&G2023 and T5$_\textsc{Large}$ for sentence lengths between 11--20 and 31--40.

    \item The F\&G2023 model performs worst on the smallest constituents and those with span lengths between 5 and 9, but ranks as the second-best model for the largest constituents.

    \item In general, BART$_\textsc{Large}$ achieves the highest performance in identifying the most frequent constituent types, although T5$_\textsc{Large}$ stands out in recognizing PP and ADJP constituents with span lengths between 3 and 6.

    \item F\&G2023 shows strength in identifying larger VP and SBAR constituents, but performs poorly on S structures (which are also relatively large). It also yields the lowest performance on shorter constituents such as ADVP and ADJP, suggesting that while the added structural information benefits long-sequence prediction, it does not help performance on shorter spans.

    \item As for BART$_\textsc{Base}$, it demonstrates strong performance on the shortest frequent constituent (ADVP), second only to BART$_\textsc{Large}$, and outperforms F\&G2023 in most cases. This indicates that incorporating a pre-trained decoder (even in a base configuration) can be more beneficial than relying solely on additional structural information in an encoder-only model.
\end{itemize}

\begin{figure}
\begin{center}
\includegraphics[width=\columnwidth]{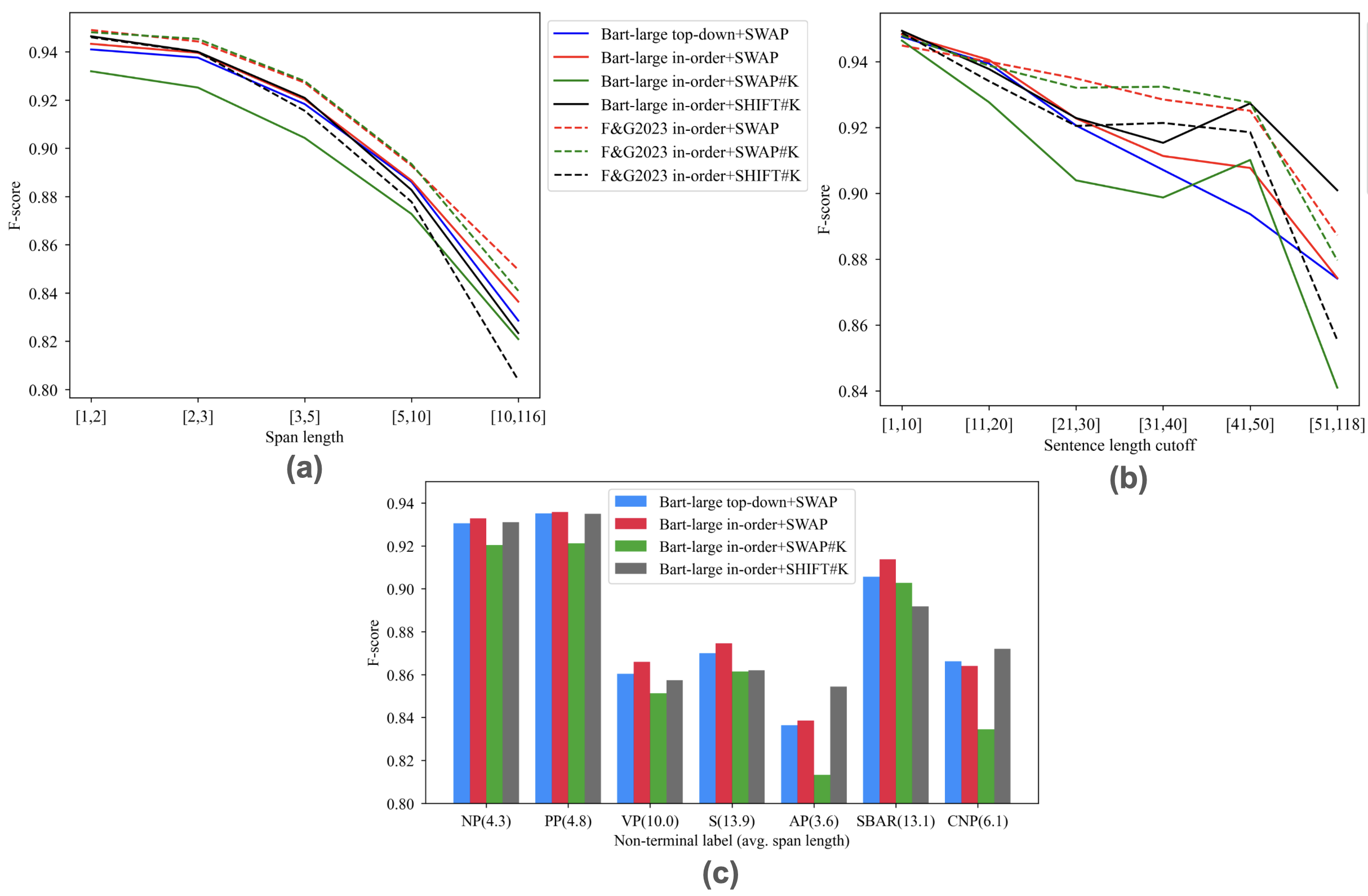}
\end{center}
\caption{F-score of lexicalized and non-lexicalized discontinuous linearizations on BART$_\textsc{Large}$/mBART and \cite{FERNANDEZGONZALEZ202343} (F\&G2023) by structural factors and sentence length.}
\label{fig:analysis3}
\end{figure}

\paragraph{Linearizations and architectures in discontinuous parsing}
Since our approach underperforms F\&G2023 on discontinuous treebanks (which require parsing longer and structurally more complex sequences), we further investigate the behavior of lexicalized and non-lexicalized discontinuous linearizations. In particular, we evaluate lexicalized variants using BART$_\textsc{Large}$/mBART, while non-lexicalized variants are assessed on the encoder-only F\&G2023 architecture. It is worth noting that F\&G2023 is initialized with RoBERTa for the English datasets and with the German-specific model GottBERT$_\textsc{Base}$ for the German treebanks, whereas our approach relies on a multilingual pre-trained encoder-decoder model. 

To this end, Figure~\ref{fig:analysis3}(a) presents the F-score for span identification across span lengths, while Figure~\ref{fig:analysis3}(b) shows performance across sentence length thresholds. For lexicalized linearizations, Figure~\ref{fig:analysis3}(c) further reports the F-score for the most frequent non-terminal labels. 
%Additionally, Figure~\ref{fig:analysis4} provides a direct comparison between each discontinuous variant of the lexicalized linearizations and its non-lexicalized counterpart in terms of identifying the most frequent constituents. 
From these results, the main observations are as follows:
\begin{itemize}
    \item The lexicalized top-down+\textsc{Swap} linearization generally underperforms the lexicalized in-order+\textsc{Swap} strategy, with the differences being particularly pronounced for the largest constituents and for sentences containing between 41 and 50 tokens.

    \item The lexicalized in-order+\textsc{Shift\#k} linearization outperforms the lexicalized in-order+\textsc{Swap} variant on short spans of fewer than 5 tokens, showing especially strong results on CNP and AP constituents. However, its performance drops considerably on larger constituents, where the in-order+\textsc{Swap} excels, particularly on VP, S, and SBAR structures. A likely explanation is that larger constituents tend to exhibit a higher degree of discontinuity, requiring the in-order+\textsc{Shift\#k} strategy to retrieve words located far from the initial token. Due to the lossy nature of this linearization, it becomes more prone to selecting incorrect words when multiple occurrences appear in the input, resulting in weaker performance on discontinuities, as shown in Table~\ref{tab:disc}.

    \item The lexicalized in-order+\textsc{Shift\#k} variant appears to outperform the in-order+\textsc{Swap} strategy on the longest sentences. This may be because the in-order+\textsc{Shift\#k} preserves the same output sequence length as the continuous in-order linearization, resulting in substantially shorter sequences than those produced when introducing explicit \textsc{Swap} actions. Therefore, the performance of the in-order+\textsc{Swap} strategy may be more susceptible to error propagation.

    \item The lexicalized in-order+\textsc{Swap\#k} variant underperforms the other two alternatives across all span and sentence length ranges. Although this strategy reduces sequence length, the introduction of a large number of \textsc{Swap\#k} tokens and the added complexity associated with learning their usage substantially harms performance. The only case where it surpasses the in-order+\textsc{Shift\#k} is in identifying SBAR constituents, which are among the largest structures. In such cases, the in-order+\textsc{Shift\#k} seems to be less effective than relying on explicit \textsc{Swap\#k} transitions.

    \item Despite being lossy, the lexicalized in-order+\textsc{Shift\#k} approach outperforms its non-lexicalized counterpart across almost all span and sentence length ranges, except for sentences containing between 31 and 40 tokens. While this strategy proves effective in our framework, the use of \textsc{Shift\#k} transitions in the F\&G2023 architecture yields the weakest results, despite incorporating additional structural information. This difference arises because our implementation leverage the pre-trained decoder to preserve the same sequence structure as the continuous in-order linearization, avoiding the introduction of additional vocabulary items associated with different values of $k$, which may be difficult for the model to learn.

    \item Non-lexicalized variants consistently outperform their lexicalized counterparts whenever \textsc{Swap} or \textsc{Swap\#k} actions are employed, with the performance gap being particularly pronounced for the latter. The structural information integrated into the pre-trained encoder-only model of \citet{FERNANDEZGONZALEZ202343} appears especially advantageous for handling \textsc{Swap} and \textsc{Swap\#k} transitions. Unlike the in-order+\textsc{Shift\#k} strategy, these transitions behave more similarly to traditional transition-based parsing operations, making explicit structural information about the stack and buffer particularly important when constructing complex discontinuous structures. Moreover, employing a German-specific model would likely further improve performance in such scenario.
    %In contrast, pre-trained encoder-decoder architectures, even without additional structural enhancements, seem better suited for linearizations such as in-order+\textsc{Shift\#k}, which more effectively leverage the decoder component. 
\end{itemize}

% \begin{figure}
% \begin{center}
% \includegraphics[width=\columnwidth]{bartVSroberta2.png}
% \end{center}
% \caption{F-score of lexicalized and non-lexicalized discontinuous linearizations on BART$_\textsc{Large}$/mBART and \cite{FERNANDEZGONZALEZ202343} (F\&G2023) when predicting the most frequent non-terminal labels (average span length shown in brackets).}
% \label{fig:analysis4}
% \end{figure}

\section{Conclusions}
\label{sec:conclusion}

In this work, we introduced the first sequence-to-sequence constituent parser built upon pre-trained encoder--decoder architectures. To better exploit the capabilities of the decoder component, we proposed a set of lexicalized tree linearization strategies derived from transition-based parsing systems for both continuous and discontinuous constituent parsing. 

We conducted an extensive evaluation across several widely used continuous and discontinuous treebanks, comparing different linearization strategies and neural architectures. The experimental results show that our approach establishes new state-of-the-art performance among sequence-to-sequence models on continuous constituent parsing benchmarks, while also achieving results competitive with strong task-specific parsers. In the discontinuous setting, our method matches the best previously reported sequence-to-sequence results on the English treebank, although it still trails behind the strongest approaches on German benchmarks, where discontinuities are typically more frequent and structurally complex.

Our analysis further highlights the importance of the interaction between tree linearizations and neural architectures. In particular, lexicalized strategies that better exploit the decoder, such as in-order+\textsc{Shift\#k}, benefit substantially from pre-trained encoder--decoder models, while transition systems relying on explicit \textsc{Swap}-based operations appear to require richer structural information to effectively model complex discontinuities.

It is important to note that our experiments rely exclusively on off-the-shelf pre-trained encoder--decoder models, without incorporating additional syntactic or structural enhancements. Consequently, our results suggest that there is still significant room for improvement. In particular, following the line of work proposed by \citet{FERNANDEZGONZALEZ202343}, future research could explore the integration of explicit structural information into encoder-decoder architectures. Combining the strengths of these models with richer stack- and buffer-aware representations may further improve discontinuous constituent parsing.

Finally, employing a German-specific pre-trained encoder-decoder model, rather than a multilingual alternative, would likely further improve performance on German treebanks.

\section*{Acknowledgments}
We acknowledge grants SCANNER-UVIGO (PID2020-113230RB-C22) funded by MICIU/AEI/10.13039/501100011033, and LATCHING-UVIGO (PID2023-147129OB-C22) funded by MICIU/AEI/10.13039/501100011033 and ERDF/EU.

\section*{CRediT authorship contribution statement}
\textbf{Daniel Fernández-González:} Conceptualization, methodology, software, validation, formal analysis, investigation, data curation, writing - original draft, writing - review \& editing, supervision, visualization. \textbf{Cristina Outeiriño Cid:} Methodology, investigation, software, data curation.

%the Spanish “Agencia Estatal de Investigación” under grants PID2019-106758GB-C31 and PID2020-113230RB-C22
%We acknowledge the European Research Council (ERC), which has funded this research under the European Union's Horizon 2020 research and innovation programme (FASTPARSE, grant agreement No 714150), ERDF/MICINN-AEI (PID2020-113230RB-C21, PID2020-113230RB-C22 and PID2023-147129OB-C22), Xunta de Galicia (ED431C 2020/11), and Centro de Investigación de Galicia ``CITIC'', funded by Xunta de Galicia and the European Union (ERDF - Galicia 2014-2020 Program), by grant ED431G 2019/01. Open Access funding provided thanks to the CRUE-CSIC agreement with Springer Nature.

% \begin{figure}[t]%% placement specifier
% \centering%% For centre alignment of image.
% \includegraphics{example-image-a}
% \caption{Figure Caption}\label{fig1}
% \end{figure}

\bibliographystyle{elsarticle-num-names}
\bibliography{anthology,main}
\end{document}